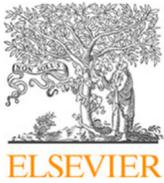
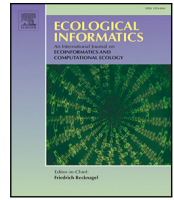

# Self-organizing maps for water quality assessment in reservoirs and lakes: A systematic literature review

Oraib Almegdadi [a,b], João Marcelino [b], Sarah Fakhreddine [c], João Manso [b], Nuno C. Marques [a,*]

[a] *NOVA LINCS, Department of Computer Science, NOVA University Lisbon, Caparica, Portugal*
[b] *Geotechnics Department, National Laboratory for Civil Engineering (LNEC), Lisbon, Portugal*
[c] *Civil and Environmental Engineering, Carnegie Mellon University, Pittsburgh, PA, USA*

## ARTICLE INFO

## ABSTRACT

Sustainable water quality underpins ecological balance and water security. Assessing and managing lakes and reservoirs is difficult due to data sparsity, heterogeneity, and nonlinear relationships among parameters. This review examines how Self-Organizing Map (SOM), the unsupervised AI technique, is applied to water quality assessment. It synthesizes research on parameter selection, spatial and temporal sampling strategies, and clustering approaches. Emphasis is placed on how SOM handles multidimensional data and uncovers hidden patterns to support effective water management. The growing availability of environmental data—from in-situ sensors, remote sensing imagery, IoT technologies, and historical records—has significantly expanded analytical opportunities in environmental monitoring. SOM has proven effective in analysing complex datasets, particularly when labelled data are limited or unavailable. It enables high-dimensional data visualization, facilitates the detection of hidden ecological patterns, and identifies critical correlations among diverse water quality indicators. This review highlights SOM's versatility in ecological assessments, trophic state classification, algal bloom monitoring, and catchment area impact evaluations. The findings offer comprehensive insights into existing methodologies, supporting future research and practical applications aimed at improving the monitoring and sustainable management of lake and reservoir ecosystems.

## Contents

* Corresponding author.
 *E-mail address:* nmn@fct.unl.pt (N.C. Marques).

1. Introduction

Water quality is fundamental to environmental management and sustainability, directly impacting both human and ecosystem health. The United Nations (UN) Sustainable Development Goals (SDGs), targeted for achievement by 2030, recognize access to clean water as a basic human need and a key factor linked to several global challenges. The UN emphasizes the need for international assistance and cooperation to reduce water pollution, enhance water quality, improve wastewater treatment, and minimize water losses (United Nations, 2012). Despite efforts by national agencies and international organizations such as the World Health Organization (WHO) to regulate and improve water quality (Boyd, 2019; World Health Organization, 2017), maintaining safe and sustainable water resources remains a global challenge. In the European Union (EU), only 40% of surface water bodies achieve good ecological status (European Environment Agency (EEA), 2021). Similarly, in the United States, data from the National Lakes Assessment (NLA) in 2012 and 2022 reveal ongoing concerns regarding nutrient pollution in lakes. In 2012, 35% of lakes had excess nitrogen and 40% had excess phosphorus. By 2022, these proportions had risen to 47% and 50%, respectively. Biological conditions also increased, with the percentage of lakes in poor biological condition rising from 31% in 2012 to 49% in 2022. These findings underscore the persistent and growing threats to aquatic ecosystems (U.S. Environmental Protection Agency, 2016, 2024).

Reservoirs and lakes are the world's primary sources of readily accessible freshwater for municipal, industrial, agricultural, and environmental uses. Increasing stressors related to ageing infrastructure, land-use change, climate variability, intensified seasonality, and rising water demand exacerbate the difficulty of protecting surface-water quality for both human and ecosystem health (Perera et al., 2021; Juracek, 2015; Asres et al., 2025). Common water-quality problems in reservoirs include sediment accumulation, turbidity increases, eutrophication, and recurrent algal blooms, all of which impair storage reliability and ecosystem function (Kondolf et al., 2014; Winton et al., 2019; Juracek, 2015).

Managing water quality in reservoirs and lakes remains challenging due to the dynamic nature of freshwater systems, where interactions among biological, chemical, and physical factors are often nonlinear, highly heterogeneous, and difficult to predict. These complexities make consistent assessment and timely intervention particularly important. Monitoring efforts are further constrained by data heterogeneity, limited spatial and temporal coverage, and environmental variability, all of which hinder compliance with national and international water-quality standards. These obstacles create increasingly complex management trade-offs and uncertainty (Wu et al., 2023), underscoring the need for continuous, reliable monitoring of water-quality dynamics in lakes and reservoirs (Miranda and Faucheux, 2022).

Advances in AI and computational tools offer promising solutions to previous challenges. AI can handle data from various sources, including in-situ measurements (Lap et al., 2023; Fernández del Castillo et al., 2024), remote sensing technologies (Tang et al., 2023; Li et al., 2024), IoT (Singh et al., 2022; Bhardwaj et al., 2022), historical archives (Ha et al., 2015; Rimet et al., 2009; Zhang et al., 2021), as well as recent approaches that integrate heterogeneous data resources (Anand et al., 2024; Xia et al., 2024; Mamun et al., 2024) to analyse patterns, optimize processes, and support predictive modelling. Various AI methods have been explored for water quality assessment, including machine learning (ML) algorithms such as $k$-means clustering, support vector machines, and neural networks (Pérez-Beltrán et al., 2024; Lowe et al., 2022).

Beyond technical complexity, water quality management is also a high-risk domain in which decisions carry direct consequences for environmental and human health, with direct implications for water security. In such contexts, human supervision together with trustworthy and responsible AI systems, is increasingly emphasized internationally to ensure that automated analyses remain reliable, transparent, and under expert oversight (NIST, 2023; European Commission, 2019; Australian Government, 2022; OECD, 2022). This orientation towards human-centred governance highlights the need for intI interpretable AI (IAI) models that promote transparency, reproducibility, and trust in environmental decision making (Gilpin et al., 2018; Rudin, 2019).

Among AI approaches, SOM introduced by Kohonen (1982b,a) and later formalized in his subsequent publications (Kohonen, 2001, 2013), offers distinct advantages for exploring and visualizing complex environmental data by preserving topological relationships between data points. SOM is a white-box model that is transparent, interpretable, and naturally aligned with human-in-the-loop analysis. It enables experts to visualize and validate emerging patterns, providing a trustworthy framework for diverse fields including environmental decision making (Wickramasinghe et al., 2021; Chon, 2011; Tang and Lu, 2022; Wandeto and Dresp-Langley, 2024; Ables et al., 2022; Costa et al., 2024; Iyer and Krishnan, 2024).

Several reviews have explored AI techniques for water quality assessment (Kalteh et al., 2008; Nikoo et al., 2025; Zhu et al., 2022; Dhapre et al., 2025; Frincu, 2025; Aderemi et al., 2025; Essamlali et al., 2024), yet they differ substantially in their objectives. Although these studies highlight valuable developments, they remain largely narrative or method-comparative, covering diverse ML algorithms as well as different water resource types with limited methodological depth and without article-level synthesis centred on unsupervised learning frameworks. For instance, Zhu et al. (2022) and Frincu (2025) presented narrative overviews of ML methods applied across different waterbody contexts, while Dhapre et al. (2025) examined AI applications solely in groundwater monitoring. Nikoo et al. (2025) offered a methodological review integrating ML, remote sensing, and statistical methods for reservoir water quality assessment, focusing on modelling techniques and data sources. Previously, Kalteh et al. (2008) provided a narrative overview of SOM usage within the broader water resources sector, integrating examples from hydrology, groundwater investigations, and water quality assessments without targeting a specific aquatic system. To our knowledge, present review provides the first systematic, article-level synthesis of SOM applications to surface-water quality in lakes and reservoirs, framed within a human-centred AI perspective. It evaluates SOM's effectiveness, interpretability, and methodological consistency through addressing the following key questions:

- **Q1. What were the main objectives for applying SOM in the reviewed studies, and what types of parameters were analysed?** This includes identifying common applications and analytical goals, as well as examining which parameter categories were most frequently used and how parameter selection varied across different study contexts.
- **Q2. What types of datasets and sampling points were used?** This assesses the nature and reliability of the data sources employed.
- **Q3. What spatial and temporal scales were considered in the studies?** This examines how SOM addressed variations in geographical coverage and study duration.

- **Q4. What were the advantages and limitations of employing SOM for water quality assessment?** This evaluates its methodological strengths and potential drawbacks
- **Q5. Which SOM libraries and computational tools were utilized?** This examines the diversity of tools, platforms, and programming environments adopted for SOM implementation, highlighting usage trends and preferences that may influence reproducibility and methodological consistency.

By addressing these questions, the review provides a structured understanding of SOM-based approaches, supporting the development of more effective water quality monitoring and management strategies.

## 2. Methodology

This systematic literature review employs a comprehensive hybrid search strategy as presented in (Fig. 1), adapted from Wohlin et al. (2022). The hybrid search strategy integrates multiple search methodologies to ensure a thorough and exhaustive review of relevant literature. Specifically, we combined a database search (DBS) with snowballing, which includes both backward snowballing (BS) and forward snowballing (FS). BS identifies new papers by searching through the reference lists of selected articles, while FS finds new papers by examining the citations of those articles (Wohlin, 2014; Wohlin et al., 2022). Snowballing was employed to ensure that no relevant articles were missed due to variations in indexing or metadata. The DBS was conducted using the Online Knowledge Library (b-on),[1] an electronic library providing access to peer-reviewed publications across multiple editor databases, including Elsevier, SpringerLink, Wiley Online Library, Taylor & Francis, and IEEE Xplore. The search query Eq. (1) was designed to capture studies related to SOM and inland water quality. The search included both journal articles and conference materials published exclusively in English. Only primary research articles were considered, while secondary sources such as reviews, editorials, and opinion pieces were excluded. No restrictions were applied to the start year of publications; however, the search was limited to studies published up to December 31, 2024. Additionally, we selected the "Apply related words" option in the b-on platform to include plurals, acronyms, and regional spellings. After removing duplicates, 260 articles were careful review to identify those focused on water sampling in reservoirs or lakes. Studies unrelated to this scope — such as those discussing groundwater, coastal areas, fossils, fish, or soil — were excluded. This screening process resulted in 27 relevant articles. An additional 8 were identified through snowballing, bringing the total to 35.

**Abstract**(("self-organizing map" OR "self-organizing maps" OR "SOM") AND

(dam OR reservoir OR wetland OR barrier OR basin OR lake)) AND

**AllText**(water OR waterbody)

(1)

## 3. Self-organizing maps (Kohonen maps)

Unsupervised learning refers to a class of AI models that discover patterns, structures, and relationships using unlabelled data (Duda et al., 2000). Among these models, SOM has emerged as a powerful tool for exploratory data analysis and visualization of high-dimensional datasets. Another widely applied unsupervised learning model, *k*-means clustering, offers a distinct approach (Hartigan and Wong, 1979). While both methods can organize data based on similarity, they use different methodologies to do so. *k*-means relies on centroid-based partitioning, whereas SOM employs a neighbourhood-based competitive learning process that preserves topological relationships within the data. Unlike k-means, which imposes hard boundaries around cluster centroids, SOM arranges similar data points close together on a structured map, allowing natural groupings to emerge. In addition, SOM outputs intuitive visual layouts that allow researchers to interpret gradual transitions and hidden structures interactively. While k-means requires predefining the number of clusters, SOM offers greater flexibility, allowing patterns to be interpreted after training—making it well suited for exploring complex environmental datasets.

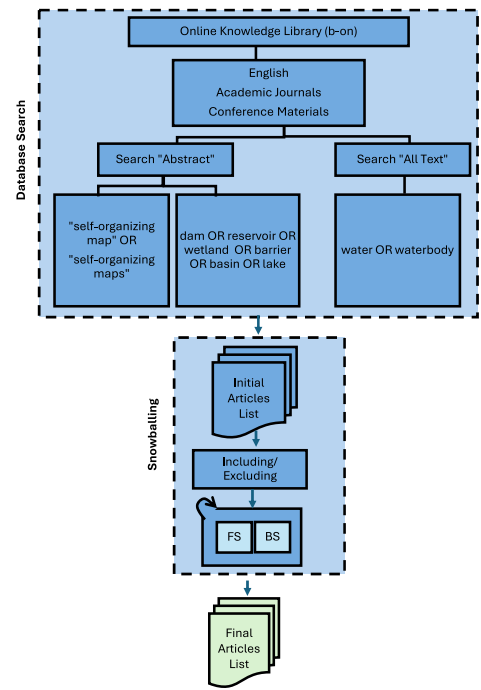

**Fig. 1.** The applied hybrid search strategy combining database queries and iterative snowballing to identify relevant literature on SOM applications in lake and reservoir water studies.

The architecture of a SOM begins with a high-dimensional input data matrix, where each observation includes multiple features. Before training, it is essential to scale all input features. Without scaling, features with larger numerical ranges may dominate the learning process, leading to distorted clustering and an unbalanced map structure. Scaling ensures that all variables contribute equally to the model, preserving the integrity of spatial relationships in the data.

During training, neurons arranged on a grid are initialized with random weights and iteratively updated through a competitive learning process. For each input vector, the Best Matching Unit (BMU)—the neuron whose weight vector most closely resembles the input—is identified. The BMU and its neighbouring neurons then adjust their weights towards the input vector, gradually organizing the grid to reflect the structure of the data.

Several tools assist in interpreting SOM outputs. The Unified Distance Matrix (U-Matrix) (Ultsch and Siemon, 1990) displays the distances between neighbouring neurons; high values reveal sparse regions, while low values indicate dense clusters. Component planes (or feature maps) visualize the distribution of individual features across the grid, typically using heatmaps to highlight gradients and trends.

In scenarios involving complex or overlapping datasets, defining clear-cut cluster boundaries can be challenging. To address this, several SOM extensions have been proposed by modifying the learning rules and grid design. The Emergent Self-Organizing Maps (ESOM) (Ultsch and Morchen, 2005) employs much larger output grids, offering finer resolution for high-dimensional structures. It enhances cluster separation using tools such as the P-Matrix (based on data density) and U*-Matrix (combining distance and density) (Ultsch, 2003a,b). Ubiquitous SOM (UbiSOM) (Silva and Marques, 2015a,b; Marques and Silva, 2023) supports continuous learning from streaming data, enabling real-time analysis without retraining from scratch. GeoSOM (Henriques et al., 2012), meanwhile, incorporates geographic coordinates into the learning process, making it especially useful in GIS-based spatial clustering. Although powerful, SOM and its variants face some limitations—including sensitivity to map size, reliance on expert judgement for interpretation, and the lack of automated methods for determining the optimal number of clusters. These concepts are illustrated in Fig. 2, which presents a SOM trained on 27 water quality parameters. Among these, the selected variables— ammonia nitrogen ($NH_3$–N), pH, and total phosphorus ($P_T$)—demonstrate clear spatial patterns across six identified clusters. Red regions on the component planes correspond to higher values of each parameter, while blue regions indicate lower concentrations, enabling intuitive interpretation of distribution gradients.

---

[1] https://www.b-on.pt/.

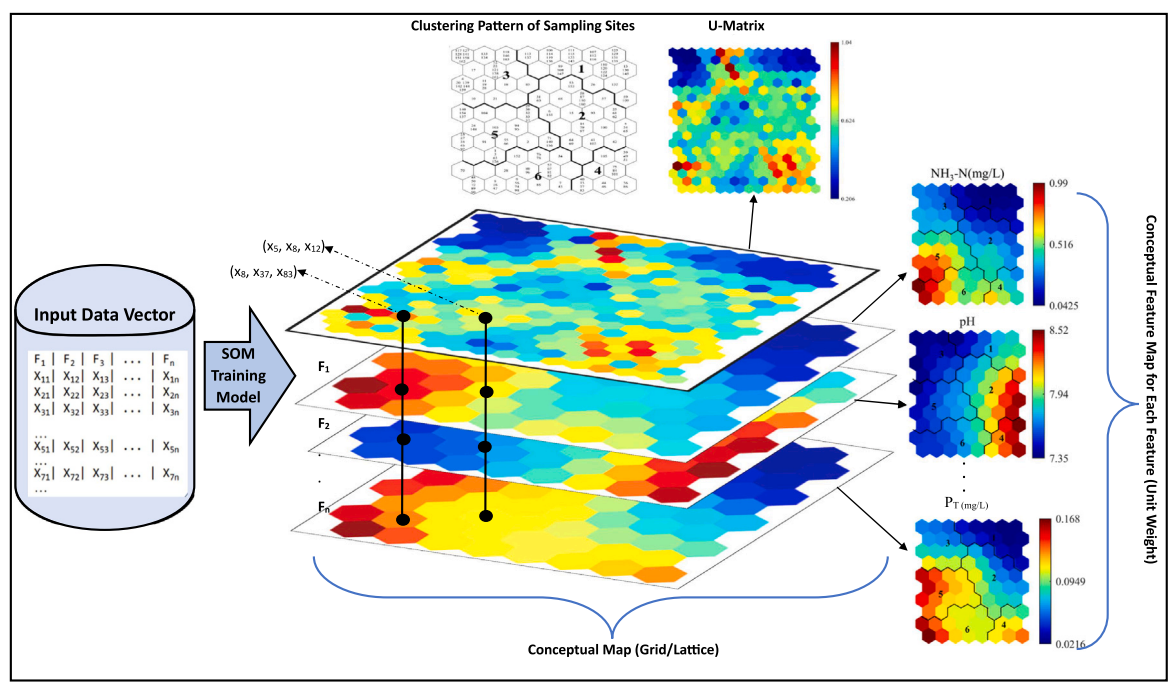

**Fig. 2.** SOM architecture. The conceptual feature maps for each parameter (denormalized values), U-matrix, and clustering pattern are adapted from Tang and Lu (2022) to illustrate the relationship between land use and water quality in a lake in China, while the figure design and structure were created by the authors.

Cluster 1 exhibits the best water quality, characterized by the lowest $NH_3$–N and $P_T$ levels and optimal pH. This cluster is associated with elevated terrains and a high proportion of forest and shrubland cover. Clusters 2 and 3 indicate moderate water quality suitable for drinking; notably, Cluster 3 occurs at higher latitudes and is linked to areas with significant forest coverage. Cluster 4 displays elevated pH and dissolved oxygen (DO) levels but only moderate $NH_3$–N and $P_T$ concentrations, correlating positively with wetland regions. Cluster 5 reflects deteriorating water quality, with elevated nutrient levels and reduced chemical oxygen demand ($COD_{Mn}$), while Cluster 6 represents the worst conditions, exhibiting the highest $NH_3$–N and $P_T$ concentrations and the lowest pH.

These spatial and environmental associations, along with detailed correlation and clustering patterns, are further explored in Tang and Lu (2022).

With the foundational concepts of SOM established, the following section reviews its practical applications in water quality studies, emphasizing how SOM has enabled the detection of environmental gradients, spatial patterns, and critical pollution zones—ultimately supporting more informed water resource management.

## 4. Analysed articles

This section presents an overview of the selected studies, highlighting the key water quality parameters assessed, the types of data analysed, and the range of SOM applications in lakes and reservoirs. A broad analysis of temporal trends and parameter usage is provided in Section 4.1, followed by a thematic classification of SOM applications in Section 4.2

### 4.1. Exploratory analysis of retrieved studies and parameters

This subsection summarizes the temporal distribution of retrieved articles, the primary parameters analysed, and general trends observed across the reviewed studies. Fig. 3 illustrates the annual distribution of retrieved articles, comparing the results of the initial search with the final selection based on the review's inclusion and exclusion criteria as detailed in the Methodology section. Although SOM was proposed in the 1980s and has proven effective in various scientific fields (Pöllä et al., 2009; Kaski et al., 1998; Oja et al., 2003), the earliest publication retrieved by our query was published in 1997 and focused on marine environments, evaluating biological and geochemical sediment samples from the back barrier tidal flats of Spiekeroog Island Kropp and Klenke (1997). The first study applied SOM specifically to water quality in lakes or reservoirs appeared in 2002, evaluating the trophic status of Fei-Tsui Reservoir (Lu and Lo, 2002). A noticeable increase in the number of initially retrieved articles was also observed starting around 2020, continuing through subsequent years. However, this surge did not result in a proportional rise in the final included studies, highlighting the importance of applying clearly defined inclusion and exclusion criteria. Much of the recent growth in publications did not align with the specific focus or methodological standards required for this review.

In addition, this review conducted a detailed quantitative synthesis of the parameters investigated across the selected studies. To facilitate interpretation, the extracted parameters were organized into seven thematic categories: (1) Nutrients and Oxygen Demand, (2) Physicochemical, (3) Biological and Ecological Indicators, (4) Metals and Metalloids, (5) Major Cations and Anions, (6) Watershed Characteristics, and (7) Others. The complete list of parameters, along with their acronyms, frequencies, and classification groups, is provided in the supplementary material. Fig. 4 presents a treemap visualization summarizing the frequency of parameters reported in the reviewed studies. Each rectangle represents an individual parameter, grouped by thematic category, while the size and colour intensity correspond to the frequency of occurrence. As notices, Nutrients and Oxygen Demand form one of the most represented groups, with parameters such as Phosphorus (P) and Nitrogen (N) frequently assessed due to their central role in nutrient enrichment. On the other hand, Physicochemical parameters are also highly prevalent, particularly water temperature ($T_W$), chlorophyll-a (Chl-a) and pH, reflecting their importance in describing the physical, chemical, and trophic status of aquatic ecosystems. Other groups appeared less consistently but contribute complementary insights into ecosystem health, contaminant monitoring, and catchment influences.

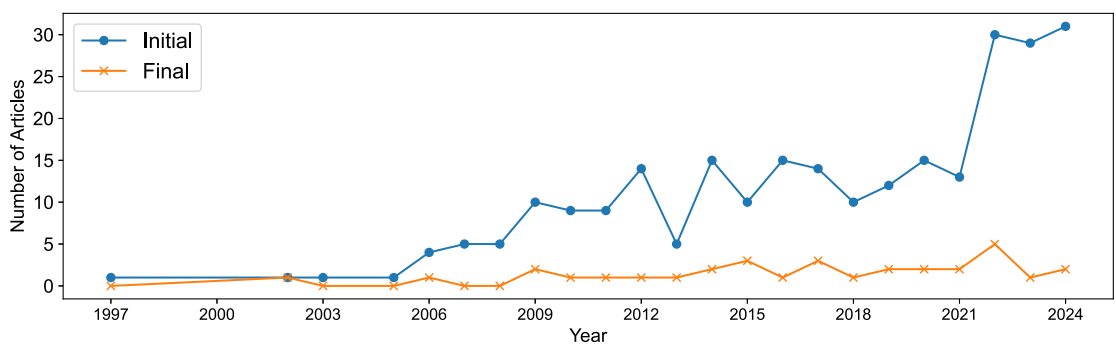

**Fig. 3.** Annual distribution of retrieved articles from the initial search compared to the final selection after applying inclusion and exclusion criteria.

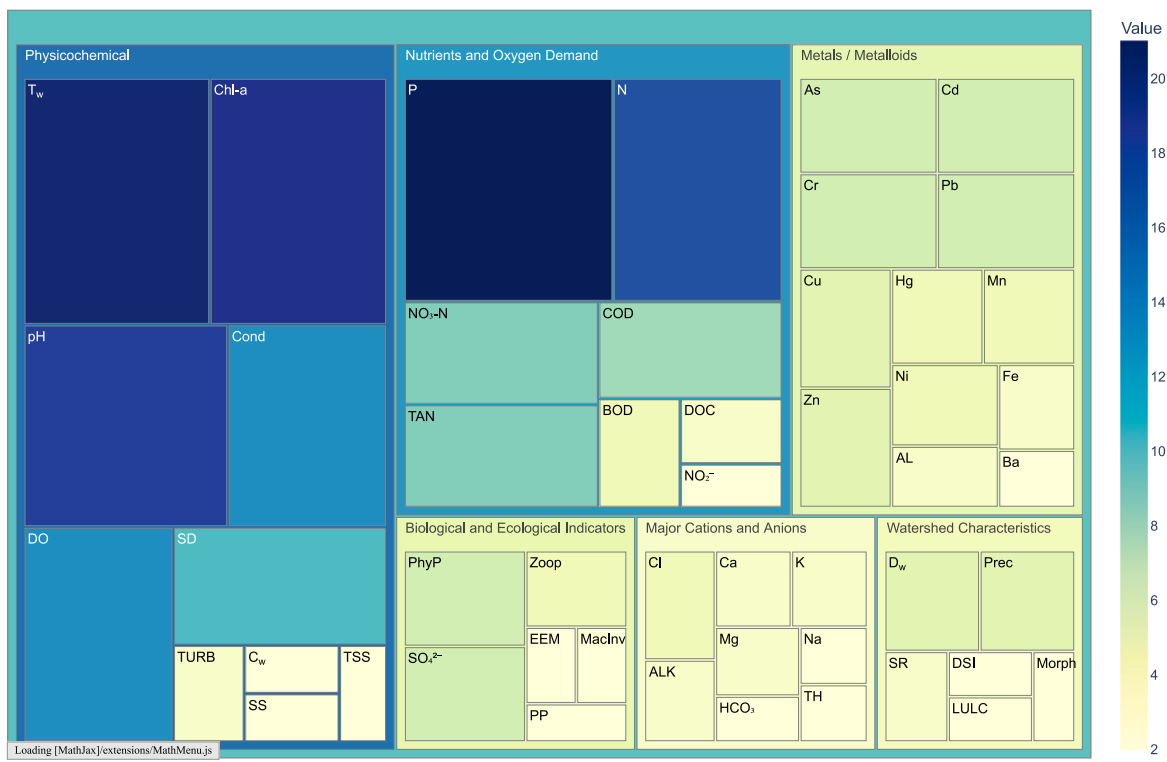

**Fig. 4.** Treemap of parameters used across the reviewed studies (Frequency > 1). Grouped by category and shaded by frequency of occurrence.

Fig. 5 illustrates the proportional representation of each parameter group across the reviewed studies, highlighting the diversity of parameters applied in different SOM applications, with the corresponding thematic domains discussed in detail in 4.2. For instance, studies investigating nutrient enrichment, algal blooms, and plankton often emphasize Biological and Ecological Indicators. In contrast, Watershed Characteristics are more prominent in catchment-scale studies but largely absent in research focusing on the monitoring of microbial indicators and pathogens, where Physicochemical parameters dominate. This distribution reflects how parameter selection aligns with specific SOM application domains.

### 4.2. Thematic applications of SOM in water quality assessment

Beyond general trends in parameter selection and study design, the reviewed literature exhibits distinct thematic patterns in the application of SOM to water quality assessment. The following subsections classify the studies according to their primary application focus, offering deeper insights into the functional roles of SOM in ecological and environmental monitoring. Due to the interdisciplinary nature of the research, thematic categorization was challenging due to the overlap in research objectives across several studies. To maintain clarity and analytical depth, each subsection presents a focused overview of studies grouped by primary application domain. Tables 1 to 5 complement the discussion by presenting structured summaries of the study area, sampling timeframe, analysed parameters, SOM configurations, and clustering outcomes.

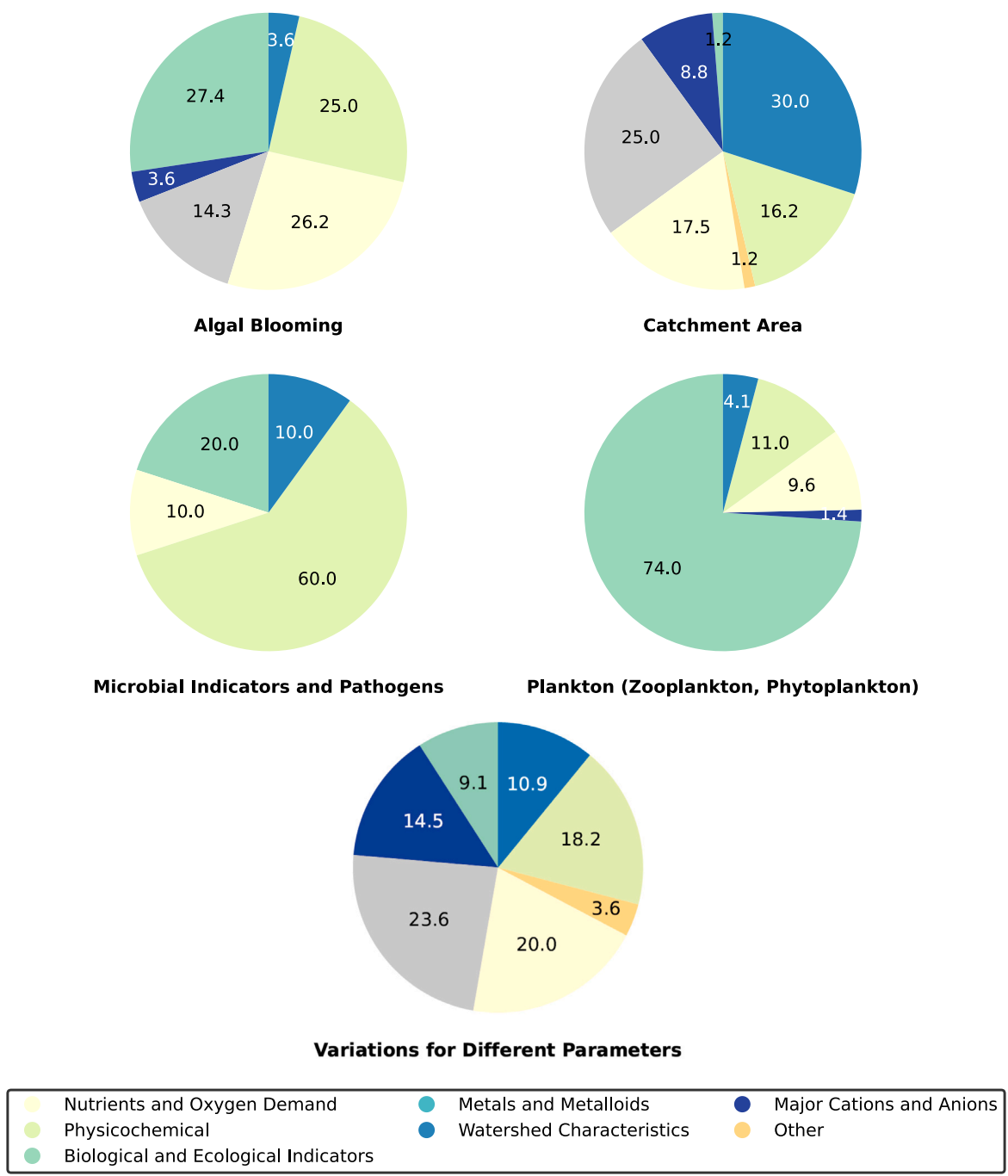

Fig. 5. Percentage composition of parameter groups across thematic applications of SOM in water quality assessment.

*4.2.1. Nutrient enrichment and algal blooms*

This section reviews studies focused on eutrophication, algal blooms, and trophic status index (TSI) in lakes and reservoirs. Eutrophication, the accumulation of nutrients in water bodies, often leads to algal blooms, reducing water quality and biodiversity. Trophic classification (e.g., oligotrophic to eutrophic) helps evaluate ecosystem productivity and condition. Continuous monitoring supports early warnings and timely management (Istvánovics, 2010; Qin et al., 2013; Carlson, 1977).

Lu and Lo (2002) addressed bias in applying TSI standards using statistical indices and proposed a model combining SOM and fuzzy theory to classify trophic levels. Voutilainen and Arvola (2017) applied three SOMs to long-term data from Lake Valkea-Kotinen, uncovering seasonal patterns, variable associations, and robustness with incomplete data. Chen et al. (2014) combined SOM and fuzzy information theory to estimate algal bloom risk, clustering environmental variables and identifying high-risk regions using fuzzy information gain. Sudriani and Sunaryani (2017) applied SOM to seasonal data from Maninjau Lake, classifying bloom risk into four categories and highlighting the influence of nutrient loading and rainfall.

Guo et al. (2020) used SOM with multivariate techniques to detect eutrophication and metal pollution patterns, identifying spatial hotspots and seasonal trends. Rimet et al. (2009) applied e-SOM to link eutrophication with phytoplankton dynamics, identifying eight diatom communities and highlighting the importance of accurate species identification.

**Table 1**
Comparison between articles discuss monitoring the algal blooming.

| Article | Study area | Samples/Time | Parameters | SOM structure | Classes |
|---|---|---|---|---|---|
| Lu and Lo (2002) | Fei-Tsui Reservoir, Taiwan | Historical data from 1987 to 1995 | Chl-a, $P_T$, SD | Input: 3*110, Train set: 90 records, test set: 20 records, Grid: 10*10 and 20*20 hexagonal | 3 |
| Chen et al. (2014) | Taihu Lake, China | Monthly 33 permanent monitoring sites from 2004 to 2010. Total: 1631 records. | Chl-a, $T_W$, CODMn, $N_T$, $P_T$ | Input: 5 × 1631, Grid: 3 × 4, hexagonal | 4 |
| Ahn et al. (2011) | Daechung reservoir, South Korea | Weekly from a floating wharf, June to November in 2001 and 2005, from July to October in 2003, and from June to October in 2004 and 2006 | $N_T$, $N_{TD}$, $N_{TP}$, $P_T$, $P_{TD}$, $P_{TP}$, $T_W$, DO, pH, TURB, Cond, SD, rainfall, DSI, cyanobacterial density | Input: 15 × 74, Grid: 9 × 6, hexagonal | Not defined |
| Voutilainen and Arvola (2017) | Lake Valkea-Kotinen, Finland | one station, weekly or biweekly from May to October, 1990–2010 (21 years). Total: 1035 records | $T_W$, DO, $C_W$, DOC, $IN_D$, $NH_4^+$, $NO_2^-$, $NO_3^-$, $PO_4^{3-}$, $N_T$, $P_T$, Chl-a, PP, Resp, Discharge, Precipitation, Cyanophyceae, Cryptophyceae, Dinophyceae, Chrysophyceae, Diatomophyceae, Raphidophyceae, Chlorophyceae, Choanoflagellata, Protozoa, Rotatoria, Cladocera, Copepoda | Input: 27 × 1035, Grid: 6 × 5, hexagona | 3 |
| Guo et al. (2020) | Gao-Bao-Shaobo Lake, China | 33 monitoring sites, April 2016 to January 2017, Seasonality | $T_W$, $D_W$, SD, Cond, DO, pH, Cl, Alk, TH, $NNH_4$, $NNO_3$, Chl-a, TSS, COD, Cu, Zn, As, Ni, Cd, Pb, Co, Cr, Mn, Fe, Al | Input: 132 × 28, Grid: 6 × 5, hexagonal | 3 |
| Rimet et al. (2009) | Geneva Lake, between France and Switzerland | One sampling point, 1974 to 2007. Total: 633 | $P_T$, $P_A$, $T_W$, $N_T$, Cond, P, $PO_4^{3-}$, pH, $SiO_2$, $NO_2$, $NO_3$, 77 taxa | database (77 taxa, 624 samples), map: 50 × 82, rectangle training: 30 | 8 |
| Hadjisolomou et al. (2018) | Megali Prespa Lake and Mikri Prespa lake, shared between Greece, Albania and Macedonia | Sampling site: Mikri Prespa: 15, Megali Prespa: 4, Seasonal, 2006–2008 | pH, DO, EC, SD, $D_W$, $T_W$, $P_T$, $IN_D$, Chl-a | Input: 105 × 9, Grid: 10 × 5, hexagonal | 5 |
| Recknagel et al. (2006) | Adjacent lakes: Veluwemeer and Wolderwijd, Netherlands | Fortnightly, 1976 to 1993. Total number of sampling points is not defined | $T_W$, SD, $NO_3$–N, $PO_4$-P, pH, Chl-a, Blue–green algae composition, Green algae, Ditom composition | Input: not defined. Grid: 15 × 11, hexagonal | 5 |
| Sudriani and Sunaryani (2017) | Maninjau Lake, Indonesia | 8 permanent sites; 2013–2014; dry and rainy seasons | Chl-a, $N_T$, $P_T$, DO, $T_W$ | Input: Not defined. Grid: 10 × 10, hexagonal | 4 |

Hadjisolomou et al. (2018) compared SOM and PCA for eutrophication analysis, with SOM outperforming due to its ability to handle nonlinear relationships and visualize correlations. SOM clearly classified samples from two lakes and provided insights into parameter interactions. Ahn et al. (2011) integrated SOM with multilayer perceptron (MLP) to assess cyanobacterial bloom dynamics. SOM visualized patterns in bloom density and environmental conditions, with temperature and dissolved nitrogen three weeks prior linked to bloom development. Recknagel et al. (2006) used SOM to distinguish two eutrophication patterns and detect seasonal behaviours, offering implications for sustainable lake conservation. Table 1 summarizes the comparative findings.

### 4.2.2. Plankton monitoring: Zooplankton and phytoplankton

Plankton, including phytoplankton (microscopic algae) and zooplankton (small drifting animals), play a central role in freshwater food webs and are sensitive indicators of ecological change. A heterogeneous dataset comprising long-term monitoring records of water quality, phytoplankton, and zooplankton was analysed using SOM (Voutilainen et al., 2012). The model effectively clustered complex multivariate data, revealing clear seasonal dynamics, interannual variations in plankton biomass, and changes in dominant species. The study highlighted the strength of SOM in visualizing high-dimensional ecological data, detecting hidden patterns, and uncovering intricate relationships between biological communities and environmental drivers in freshwater ecosystems. Ha et al. (2015) aimed to identify relationships between dominant zooplankton species and environmental conditions during different study periods, pre-, during- and post-biomanipulation. The SOM successfully categorized the limnological characteristics of the lake into distinct features and provided insights into the temporal changes in zooplankton communities and their interactions with the evolving environmental conditions throughout the study period. Banerjee et al. (2022) applied SOM to investigate the relationship between zooplankton distribution and environmental variables in a tropical reservoir with a diverse trophic structure. To enhance interpretability and computational efficiency, the dataset was divided into two subsets: ENVS (environmental variables) and ZOOP (zooplankton metrics). The SOM analysis successfully revealed spatial and temporal patterns in zooplankton communities, highlighting key environmental drivers and supported the utility of SOM in ecological pattern recognition. In Oh et al. (2007) the temporal patterns of cyanobacteria were studied using a combination of SOM and the MLP. The SOM was initially used to cluster the phytoplankton communities based on their algal composition (Cyanophyceae, Chlorophyceae, Bacillariophyceae, and others). Then the MLP was applied to identify the major environmental factors causing the abundance of phytoplankton in the clustered communities. Table 2 illustrates comparisons between these articles.

Table 2
Comparison between articles discuss monitoring Plankton (Zooplankton, Phytoplankton).

| Article | Study area | Sampling details | Parameters | Input/Grid | Classes |
|---|---|---|---|---|---|
| Ha et al. (2015) | Shirakaba Lake, Japan | Weekly sampling from a floating wharf: June to November in 2001 and 2005; July to October in 2003; and June to October in 2004 and 2006 | pH, DO, $T_W$, SD, Chl-a, $N_T$, $P_T$, Biomass of zooplankton species: Bosmina longirostris, Daphnia galeata, Holopedium gibberum, Leptodora kindtii, Cyclops spp., Eodiaptomus japonicus, Nauplius, Filinia longiseta, Keratella cochlearis, Polyarthra spp., Trichocerca spp. | 12 218/97, hexagonal | 5 |
| Banerjee et al. (2022) | Bakreswar reservoir, India | 3 stations, 4 samples/year (seasonally), 2 years | $T_A$, $T_W$, SR, $N_T$, P, SAL, Alk, BOD, N, P, Chl-a, Cond, TH, $N_{TP}$, DO, density counts per litre of: Cladocera, Copepoda, Ostracoda, and Rotifera | ENVS Input: 144 × 14, Grid: 10 × 10, hexagonal. ZOOP: Input: 144 × 4, Grid: 10 × 10, hexagonal | 3 |
| Oh et al. (2007) | Daechung reservoir, Korea | Weekly sampling: Apr–Oct 1999, Jun–Nov 2001, July–Oct 2003 | $T_W$, pH, Cond, DO, SD, Chl-a, $N_T$, $P_T$, Precipitation, DSI, %Cyanophyceae, %Chlorophyceae, %Bacillariophyc, %others | Input: 39 × 4, Grid: 6 × 4, hexagonal | 4 |
| Voutilainen et al. (2012) | Pyhäselkä Lake, Finland | 5 sampling points, 1987–2009 | $T_W$, Precipitation, $P_T$, $N_T$, Chl-a, $C_W$, pH, EC, Phytoplankton production, Phytoplankton biomass, Chlorophyta, Chrysophyceae, Cryptophyta, Diatomophyceae, Cyanophyta, Rotatoria biomass, Asplancha spp, Cladocera biomass, Bosmina spp., Chydorus spp., Holopedium gibberum, Polyphemus pediculus, Daphnia cristata, Limnosida frontosa, Leptodora kindti, Copepoda biomass, Eudiaptomus spp., Cyclops spp., Meso-Thermocyclops spp., Heterocope appendiculata, Limnocalanus macrurus, Eurytemora lacustris | Input: 163 × 32, Grid: 8 × 8, hexagonal | 7 |

*4.2.3. Identifying the impact of catchment area characteristics*

Catchment areas critically shaped the chemical and biological conditions of water bodies through runoff composition, nutrient loading, and land-use pollution, influencing water quality management. Ding et al. (2019) addressed the relationship between water quality measurements and land use indicators across various spatiotemporal scales by analysing the influence of topographical variations during different seasons on the reservoirs. Wang et al. (2020) aimed to conduct source apportionment and risk assessment of trace metals in urban aquatic environments, while accounting for uncertainty. SOM was used to categorized trace metals into distinct clusters, revealing their potential origins. Combined with Positive Matrix Factorization (PMF), the analysis identified dominant anthropogenic sources including agriculture, industry, vehicles, and natural geogenic inputs. This approach enhanced the understanding of spatial pollution patterns and supported informed water and land-use management. Melo et al. (2019) evaluated the effectiveness of SOM in analysing biogeochemical processes and the influence of land use on reservoir water quality. Seven sampling sites were examined, with surrounding land use classified into five categories using supervised Landsat 8 data, refined with RapidEye imagery.

SOM revealed eutrophication near urban and pasture areas, highlighting their value in identifying critical zones to inform effective management and conservation policies linking land use with water quality. Based on physicochemical variables, SOM classified the reservoirs into six clusters, ranging from least to most polluted (Park et al., 2014). Similarly, SOM effectively illustrated the spatial distribution of monomethylmercury in reservoirs, uncovering complex interactions between hydrologic and chemical factors (Noh et al., 2016). The clustering analysis revealed spatial variance and inverse relationships within the dataset, providing valuable insights for identifying contamination patterns and supporting environmental management strategies. Moreover, SOM enabled the comprehensive identification of inverse relationships, assessment of vulnerabilities, and detection of spatial patterning, contributing further to environmental analysis and decision making. In a related study, Wu et al. (2024) investigated the relationship between land use and water quality in Lake Dianchi, China, over a 10-year period. By combining remote sensing and in-situ surface water sampling, the study analysed the effects of urbanization, agriculture, and land development on pollution levels. SOM clustering revealed that pollution was highest near urban and semi-urban transition zones, emphasizing the need for multi-dimensional governance to manage pollution sources and enhance water quality in lakeshore environments. Table 3 illustrates comparisons between these articles.

*4.2.4. Dynamic variations across water quality parameters*

This subsection reviews studies that applied SOM to analyse spatial and temporal variations in water quality across physicochemical, biological, and spectroscopic parameters. Simeonova et al. (2010) classified Bulgarian high-mountain lakes based on pH, conductivity, and major ions, revealing spatial patterns and geochemical groupings. Farmaki et al. (2013) compared MLP, RBF, and SOM to evaluate water quality in Athens reservoirs, with SOM offering superior interpretation. Similarly, Li et al. (2015) highlighted SOM's value in supporting water pollution control through cluster-based temporal analysis, while Zhang et al. (2015) used SOM to identify relationships between phosphorus and other variables for real-time monitoring. Li et al. (2017) grouped 27 lakes by trophic status and examined Chl-a variation, stressing the importance of nonlinear interactions. Yang et al. (2017) and Mohammed et al. (2022) both used SOM to visualize heavy metal distributions, identifying pollution hotspots and contamination sources.

Dou et al. (2022) clustered sites in Lianhuan Lake using macroinvertebrate data, showing SOM's capability in revealing ecological gradients, even when trained on biological variables alone. Uniquely, Jin

Table 3

Comparison between articles discuss the impact of catchment area.

| Article | Study area | Samples/Time | Parameters | SOM structure | Classes |
|---|---|---|---|---|---|
| Ding et al. (2019) | 61 Reservoirs, Zhejiang Province, China | Monthly 2015–2016 | pH, DO, CODMn, $P_T$, $N_T$, $NH_3$–N | Input: 31 × 6, Grid: 7 × 6, hexagonal | Not defined |
| Wang et al. (2020) | Wanshan Lake, China | 30 samples/ April. 2019 | Cd, Mn, Ni, Cu, Zn, As, Cr, Hg, Pb | Input: 30 × 9, Grid: 8 × 4, hexagonal | 3/4 |
| Melo et al. (2019) | Itupararanga Reservoir, Brazil | 7 sampling points, 2 depths, 2016 (Dec), 2017 (Mar, Aug, Oct, Dec), 2018 (Mar, Nov) | ALK, CLI, DO, EC, Eh, pH, $SO_4^{2-}$, $T_W$, TDS, TURB, DOC, TOC, $N_T$, $P_T$, $AL_D$, $AL_T$, $Ba_T$, $Ba_D$, $Ca_D$, $Fe_D$, $Fe_T$, $K_D$, $Mg_D$, $Mn_D$, $Mn_T$, Chl-a, $NO_3$, $NH_4$, $NO_4^3$, %Agriculture, %forest, %Field, %silviculture, %urban | Input: 98 × 31, Grid: 12 × 12, hexagonal | 17 |
| Park et al. (2014) | 302 reservoirs, Korea | Time not specified | DO, COD, TSS, $N_T$, $P_T$, Chl-a, Altitude, Bank height, Bank width, Circumference, Reservoir length, Surface area, Pondage, %Urban, %Forest, %Paddy field, %Dry field, %Grassland; | Input: 302 × 6, Grid: 12 × 7, hexagonal | 6 |
| Noh et al. (2016) | 7 reservoirs, South Korea | 33 samples, Jun–Sep 2013, 2014 | $T_W$, Cond, pH, SPM, Chl-a, $SO_4^{2-}$, $NO_3^-$, DOC, precipitation (3/30 days), Water residence time, Reservoir area, Reservoir volume, Maximum depth, Reservoir depth, Hg, MeHg, Log-transformed MeHg concentration | Input: 33 × 19, Grid: 5 × 4, hexagonal | 5 |
| Wu et al. (2024) | Lake Dianchi, China | 100 sampling points, 6 time points (2014–2022) | N, $P_T$, S, Cd, Cr(VI), Pb, COD, VP, AS | Input: 9 × 600, Grid: 10 × 9, hexagonal | 3 |

et al. (2023) and Song et al. (2024) applied SOM to EEM fluorescence data to characterize DOM in lakes. While Jin et al. (2023) compared lakes influenced by agriculture versus urban runoff, Song et al. (2024) identified spectral patterns linked to land use. Both studies underscored SOM robustness to noise and suitability for high-dimensional spectroscopic data.

Finally, Chrobak et al. (2021) applied SOM to support expert systems in assessing lake conditions in Poland using a nine-dimensional ecological dataset. SOM outputs aligned with the Water Framework Directive and were combined with expert judgement to enhance interpretability and guide ecological interventions. A comparative summary of these articles is presented in Table 4.

#### 4.2.5. Monitoring of microbial indicators and pathogens

Several studies applied SOM to monitor microbial and biological indicators in aquatic environments, including a hybrid approach by Ahmad et al. (2009) presented a hybrid approach using SOM and Fuzzy Logic to predict coliform presence. SOM generated fuzzy membership rules from environmental variables, enhancing a module that predicted real-time coliform levels with 90.5% accuracy. In Bertone et al. (2019) SOM was utilized to visually explore correlations between multiple variables and to understand the relationships within the dataset. The SOM helped in identifying patterns and trends related to E. coli blooms, such as the influence of warm water, dry catchments, algal blooms, and nutrient availability. In another study, Son et al. (2021) used SOM to examine invertebrate distribution in relation to aquatic macrophytes and environmental conditions in a shallow reservoir, revealing that specific groups preferred different macrophyte species and highlighting the role of vegetation in shaping biological community structure. Similar, Lee et al. (2022) used SOM to cluster taxa based on 12 dichotomous traits, revealing patterns in trait associations across 20 South Korean reservoirs. The SOM model visualized community-level functional diversity along environmental gradients, enhancing interpretation of how macroinvertebrate traits responded to spatial habitat variability for ecosystem assessment and biomonitoring. Table 5 illustrates comparisons between these articles.

### 5. Discussion

In this section, we explore key methodological aspects of applying SOM to water quality assessment. We examine clustering techniques and their interpretability, spatial–temporal data sampling strategies, commonly used tools for implementing SOM, and its role in predictive modelling. By drawing on insights from the reviewed studies, we highlight both the strengths and limitations of current approaches, as well as broader implications for environmental monitoring and decision making. Understanding these aspects is essential not only for enhancing methodological robustness but also for translating SOM-based insights into actionable water management strategies.

#### 5.1. Subjective and objective clustering : Insights from reviewed studies

SOM is frequently employed for clustering, often by assigning each data point to its BMU. However, this approach overlooks SOM's original purpose as a topology-preserving visualization and vector quantization method that projects high-dimensional data onto a low-dimensional grid while maintaining the spatial relationships among inputs (Kohonen, 1982b,a). Clusters derived from SOM are commonly interpreted through the U-Matrix, which visualizes the distances between adjacent neurons. High values on the U-Matrix indicate sparse regions (i.e., large distances between neurons), while low values indicate dense regions (Ultsch and Siemon, 1990). This method is helpful for exploratory analysis but introduces subjectivity, as the delineation of clusters depends on the colour scale and the interpreter's judgement.

To contextualize SOM-based clustering, it is useful to contrast it with other techniques. The *k*-means algorithm explicitly partitions data into *k* clusters by minimizing within-cluster variance, resulting in distinct clusters defined by Voronoi boundaries (MacQueen, 1967). PCA reduces dimensionality by projecting data onto orthogonal axes that capture maximal variance (Hotelling, 1933; Jolliffe, 2002). Fuzzy clustering algorithms, such as Fuzzy C-Means (FCM), assign membership probabilities based on distance metrics, allowing data points to belong to multiple clusters simultaneously (Bezdek, 1981; Dunn, 1973). t-Distributed Stochastic Neighbour Embedding (t-SNE), while useful for visualizing local structure in high-dimensional data, lacks a direct clustering criterion (Maaten and Hinton, 2008; Van Der Maaten, 2014; Wang et al., 2020) While these methods offer objective or probabilistic frameworks, SOM's emergent groupings should be interpreted as topological regions rather than definitive clusters. To avoid mischaracterization, it is important that U-Matrix zones be described as visual structures, not algorithmically derived clusters. Despite these limitations, many reviewed studies relied on SOM for clustering, and attempted to overcome subjectivity through quantitative validation methods. Son et al. (2021) examined the Euclidean distances between neurons on the U-Matrix to identify transitions between clusters. In scenarios with overlapping datasets, such as theirs, visual delineation of clusters was noted as particularly challenging. The authors acknowledged this limitation and used the U-Matrix to observe transition zones, reinforcing the value of SOM for visual exploration while also highlighting its interpretative limitations. To improve boundary clarity in such cases, the U*-Matrix from ESOM has been proposed as a refined visualization tool (Ultsch, 2003a; Ultsch and Morchen, 2005; Ultsch, 2007), though interpretation remains reliant on expert insight. To enhance interpretability and objectivity, Banerjee et al. (2022) used the Elbow method to determine the optimal number of clusters. Chrobak et al. (2021) performed a comparative analysis using both the Elbow and Silhouette methods (Rousseeuw, 1987). Interestingly, the Elbow method indicated three clusters, while the Silhouette method suggested two. Despite these findings, the authors ultimately selected five clusters, aligned with pre-existing ecological classifications of lake states, illustrating how domain knowledge can override statistical recommendations in environmental applications. Studies (Wang et al., 2020; Li et al., 2017; Ha et al., 2015) employed the Davies–Bouldin Index (Davies and Bouldin, 1979) to evaluate clustering compactness and separation. On the other hand, Uriarte and Díaz Martín (2005) and Ding et al. (2019) used the Topographic Error (TE) and Quantization Error (QE) to evaluate the structural quality of the trained SOM maps. Liu et al. (2022) applied the Calinski–Harabasz Index (Caliński and Harabasz, 1974), which is particularly effective for identifying the structure of multivariate datasets. Additionally, some articles applied the *k*-means algorithm to cluster the trained SOM output and classify the sampling points into clusters such as in Yang et al. (2017) and Dou et al. (2022). The articles (Ahn et al., 2011; Chen et al., 2014) did not provide a clear explain the method used for defining the number of classes or provide a clear explanation of the clustering division.

Table 4
Comparison between articles discuss the variations in water quality for different parameters.

| Article | Study area | Samples/Time | Parameters | SOM structure | Classes |
|---|---|---|---|---|---|
| Simeonova et al. (2010) | 39 lakes in Rila Mountain, Bulgaria | May. 2003–October. 2004 | pH, DM, Cond, Ca, Mg, Na, K, $HCO_3$, $SO_4$, Cl, $NO_3$ | Input: 39 × 11, Grid: 8 × 6, hexagonal | 6 |
| Farmaki et al. (2013) | 3 water reservoirs, Greece | 89 samples, Oct. 2006–Apr. 2007 | Fe, B, Al, V, Cr, Mn, Ni, Cu, Zn, As, Cd, Ba, Pb | Input: 89 × 13, Grid: 6 × 8, hexagonal | 3 |
| Li et al. (2015) | Lake Taihu, China | 8 monitoring sites, Monthly 2002–2006 | $D_W$, $T_W$, SD, SS, DO, $NH_4$–N, $NO_2$–N, $NO_3$–N, $PO_4$, $SO_4$, $P_T$, $N_T$, Chl-a, Pheo | Input: 672 × 14, Grid: 9 × 15, hexagonal | 10 |
| Zhang et al. (2015) | Honghu Lake | Jan. 2010: 32 samples; Oct. 2010: 33 samples, July. 2011: 41 samples. Sampling sites spread all over the lake without deifying exact location | SD, $T_W$, pH, Cond, DO, $P_T$ | Input: 106 × 6, Grid: 12 × 6, hexagonal | Not defined |
| Li et al. (2017) | 27 lakes, China | At least three samples/year. 2009–2011 | $T_W$, pH, SD, DO, CODMn, BOD, $NH_3$–N, PET, $N_T$, $P_T$, Chl-a, volatile phenol, Hg, Pb, Cu, zn, fluoride, Se, As, Cd, Cr, CN, ASAA, $S^{2-}$ | Input: 149 × 24, Grid: 8 × 10, hexagonal | 4 |
| Yang et al. (2017) | 4 lakes, Kenya | Sampling points: Lake Naivasha: 30, Lake Elementaita: 20, Lake Nakuru: 20, and Lake Bogoria: 30. Jan. 2014 | Cr, Ni, Cu, Zn, As, Pb, Cd, Hg | Input: 100 × 8, Grid: 8 × 6, hexagonal | 3 |
| Dou et al. (2022) | 13 lakes, China | Spring (Jun), summer (Aug), and autumn (Oct) of 2020 | $D_W$, $T_W$, pH, COND, CODMn, $NH_3$–N, $NO_3$–N, $NO_2$–N, Chl-a, SS, $P_T$, $N_T$, Macroinvertebrates Abundance | Input: 74 × 14, Grid: 7 × 7, hexagonal | 5 |
| Mohammed et al. (2022) | Boudaroua Lake, Morocco | 5 sampling points, Jan. 2020, Oct. 2019, Apr. 2021, Jun. 2019 | $T_W$, pH, EC, BOD, COD, $NH_4$, $NO_3$, Cl, $HCO_3$, $SO_4$, Na, Ca, Mg, K | Input: 20 × 14, Grid: 8 × 4, hexagonal | 4 |
| Jin et al. (2023) | Gaotang Lake (GT) and Yaogao Reservoir (YG), China | 43 samples (GT: 23, YG: 20) – Jan 2022 | EEM | Input: dimension is not clear, Grid: 12 × 3, hexagonal | 2 |
| Song et al. (2024) | TX and MC Lakes, China | 24 samples (Dec 2022, May 2023) | EEM | Input: dimension is not clear, Grid: 8 × 3, hexagonal | 2 |
| Chrobak et al. (2021) | 496 lakes in Poland | 2010–2015, surface water samples under national ecological monitoring | ESMI, PMPL, IOJ, Chl-a, N, P, Phytoplankton, Visibility | Input: 9 × 496, Grid: 10 × 10 hexagonal | 5 |

Table 5
Comparison between articles discuss the monitoring of microbial indicators and pathogens.

| Article | Study area | Sampling details | Parameters | Input/Grid | Classes |
|---|---|---|---|---|---|
| Ahmad et al. (2009) | Putrajaya Lakes, Malaysia | 5 years, number of sampling points not defined | pH, $T_W$, BOD, TC | Input: not defined, Grid: 16 × 13, hexagonal | 3 |
| Bertone et al. (2019) | Five reservoirs, Australia | Neither number of samples nor study period defined | Not defined | Not defined | Not defined |
| Son et al. (2021) | Jangcheok Reservoir, South Korea | 50 sampling points, 3 times during spring (May–Jun) | $D_W$, pH, Cond, TURB, Chl-a | Input: 150 × 24, Grid: 7 × 5, hexagonal | 5 |
| Lee et al. (2022) | 20 reservoirs, South Korea | 2 benthic stations per reservoir, twice per year (spring and autumn) from 2009–2017; | Macroinvertebrate taxa features: number of generations per year (voltinism), adult lifespan, adult size to maturity, functional feeding groups | Input: 85 × 12, Grid: 5 × 8, hexagonal | 3 |

In summary, while SOM serves as a powerful tool for visualizing complex environmental data, the clustering derived from it is often subjective unless complemented by objective methods. The reviewed literature demonstrates an increasing trend towards integrating quantitative validation metrics and algorithmic techniques, which enhances the robustness and transparency of clustering outcomes in SOM-based water quality analysis.

*5.2. Data sampling strategies*

The reviewed studies employed three main sampling strategies: spatial, temporal, and spatiotemporal. Spatial sampling involved collecting data from multiple locations at a single point in time to detect spatial heterogeneity and identify localized pollution sources. This approach involved sampling multiple sites within a single water body, as demonstrated by Wang et al. (2020) in Wanshan Lake, or sampling across different reservoirs simultaneously, as applied by Park et al. (2014) in South Korea. Temporal sampling, in contrast, focused on repeated measurements at the same location over time, enabling the capture of seasonal variations, long-term trends, and sudden environmental events. For example, Rimet et al. (2009) monitored Geneva Lake from 1974 to 2007 through a long-term temporal sampling strategy. Spatiotemporal sampling combined both dimensions to provide a more comprehensive understanding of dynamic patterns, as demonstrated by Simeonova et al. (2010) in the monitoring of 39 lakes in the Rila Mountains, Bulgaria. The water bodies analysed in these studies also varied considerably in physical size, ranging from small lakes such as Lake Valkea-Kotinen in Finland, where a single weekly sampling location was considered sufficient for representative monitoring (Voutilainen and Arvola, 2017), to large systems like Lake Taihu in China, where extensive spatiotemporal sampling was necessary to capture significant heterogeneity, as demonstrated by Li et al. (2015). These examples highlighted how the design of the sampling strategy is closely tied to the size and complexity of the water body under study.

Overall, spatiotemporal sampling emerged as the most dominant approach in the reviewed studies, applied in approximately 71% of the cases, followed by temporal sampling 20% and spatial sampling 9%. Fig. 6 and Table 6 illustrate the distribution of articles across these categories.

The adaptability of SOM demonstrated strong effectiveness across varied sampling strategies. Their ability to manage and visualize high-dimensional, irregularly spaced datasets enabled researchers to extract meaningful patterns across both spatial and temporal domains. In studies utilizing extensive spatiotemporal sampling, SOM facilitated the interpretation of spatial variability and temporal trends, making them highly suitable for addressing the complex dynamics of waterbody monitoring. Furthermore, their clustering capabilities, combined with topological preservation, allowed for intuitive visualization of evolving water quality conditions, thereby strengthening exploratory analysis and supporting more informed management decisions.

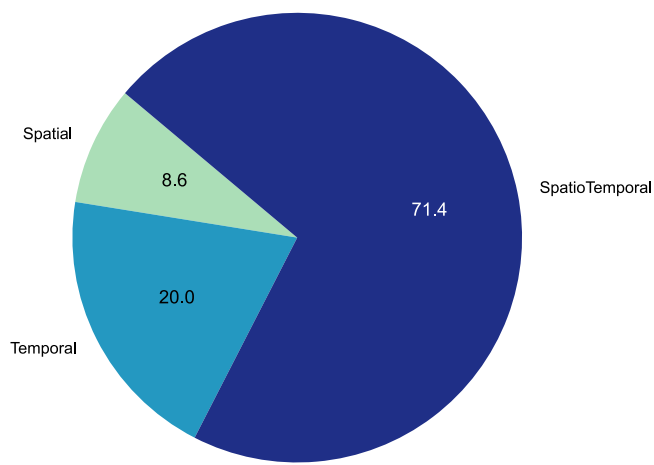

**Fig. 6.** Percentages of sampling methodologies applied in the reviewed articles.

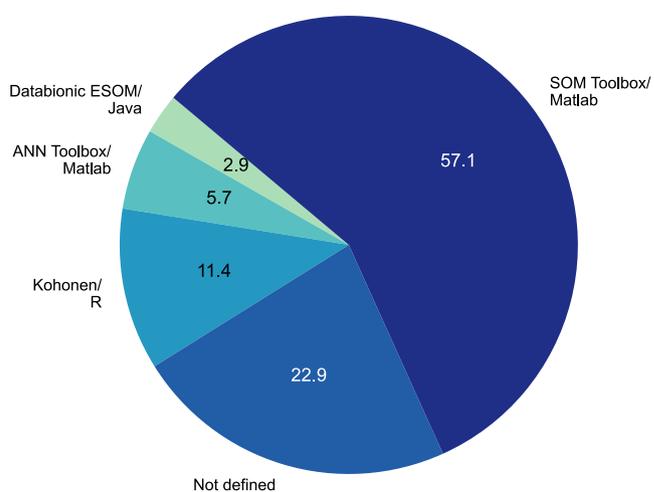

**Fig. 7.** Percentages of libraries and tools utilized for SOM implementation and analysis.

*5.3. Tools for implementing SOM*

This section explores the technological trends in implementing SOM, highlighting the diversity of libraries, platforms, and programming environments used across the reviewed studies. As shown in Table 7 and Fig. 7, the SOM Toolbox for MATLAB, developed by Helsinki University of Technology, was the most widely adopted tool, appearing in approximately 57% of the reviewed studies. The Kohonen package for R ranked second, used in 11.4% of cases, offering an open-source alternative particularly appealing to researchers within the R environment. The ANN Toolbox for MATLAB was applied in 5.7% of studies, while Databionic ESOM for Java represented for 3%. Notably, 23% of the studies did not specify the tool or platform used. Overall, MATLAB-based environments (SOM Toolbox and ANN Toolbox combined) accounted for approximately 63% of the reported implementations, reflecting the platform's widespread adoption, established computational capabilities, and the early availability of dedicated SOM libraries.

*5.4. Prediction, interpretability, and deployment*

Although SOM is primarily designed for exploratory analysis and visualization, their predictive and interpretable potential have been explored in several studies. For instance, Ahn et al. (2011) reported that SOM exhibited limited variation and low predictability for cyanobacterial bloom peaks, functioning more as a "lookup table" that correlated environmental variables with cyanobacterial density values. However, the model's predictive performance was hindered by data limitations and an averaging effect, resulting in lower accuracy compared to MLP, particularly for forecasting peak events. Similarly, Zhang et al. (2015) found that while SOM outperformed multivariable linear regression models in predicting $P_T$ concentrations, challenges remained in fully capturing complex environmental dynamics.

**Table 6**
Categorization of articles based on data sampling methodologies.

| Category | Articles |
| --- | --- |
| Spatial | Park et al. (2014), Yang et al. (2017) and Wang et al. (2020) |
| Temporal | Lu and Lo (2002), Oh et al. (2007), Rimet et al. (2009), Ahn et al. (2011), Ha et al. (2015), Voutilainen and Arvola (2017) and Lee et al. (2022) |
| Spatiotemporal | Recknagel et al. (2006), Ahmad et al. (2009), Simeonova et al. (2010), Voutilainen et al. (2012), Farmaki et al. (2013), Chen et al. (2014), Li et al. (2015), Zhang et al. (2015), Noh et al. (2016), Li et al. (2017), Sudriani and Sunaryani (2017), Hadjisolomou et al. (2018), Ding et al. (2019), Melo et al. (2019), Bertone et al. (2019), Guo et al. (2020), Chrobak et al. (2021), Son et al. (2021), Mohammed et al. (2022), Dou et al. (2022), Banerjee et al. (2022), Liu et al. (2022), Jin et al. (2023), Song et al. (2024) and Wu et al. (2024) |

**Table 7**
Technological preferences and trends in SOM research.

| Library name articles | |
| --- | --- |
| SOM Toolbox(MATLAB) | Recknagel et al. (2006), Ahn et al. (2011), Voutilainen et al. (2012), Farmaki et al. (2013), Park et al. (2014), Zhang et al. (2015), Ha et al. (2015), Noh et al. (2016), Yang et al. (2017), Hadjisolomou et al. (2018), Ding et al. (2019), Melo et al. (2019), Mohammed et al. (2022), Voutilainen and Arvola (2017), Li et al. (2015), Son et al. (2021), Wang et al. (2020), Lee et al. (2022), Jin et al. (2023) and Song et al. (2024) |
| Kohonen Library | Chrobak et al. (2021), Banerjee et al. (2022), Liu et al. (2022) and Wu et al. (2024) |
| ANN ToolBox(MATLAB) | Guo et al. (2020) and Dou et al. (2022) |
| Databionic ESOM Tools(Java) | Rimet et al. (2009) |
| Not defined | Lu and Lo (2002), Simeonova et al. (2010), Chen et al. (2014), Oh et al. (2007), Ahmad et al. (2009), Sudriani and Sunaryani (2017), Li et al. (2017) and Bertone et al. (2019) |

It is important to note that several extensions of SOM have been developed to enhance its predictive capability. Learning Vector Quantization (LVQ) (Kohonen, 1995) fine-tunes the SOM output by incorporating labelled training data, enabling classification and prediction based on supervised learning principles. The Supervised Self-Organizing Map (SSOM) (Van Hulle, 2000) further integrates class label information directly into the training phase, improving discrimination during map creation. The Recurrent Self-Organizing Map (RSOM) (Hammer and Villmann, 2002) and Temporal Kohonen Map (TKM) (Chappell et al., 1996) extend SOM for temporal sequence processing and time-series prediction. None of the reviewed studies explicitly employed these supervised or recurrent variants, possibly reflecting the exploratory focus of the analyses or the adequacy of combining SOM with downstream predictive models.

From a broader standpoint, IAI emphasizes transparency by enabling users to understand a model's internal logic rather than relying on output justifications, and it ensures that humans remain actively involved in interpreting model behaviour. Explainable AI (XAI) methods such as SHAP and LIME complement this perspective by providing post-hoc explanations for models whose internal structure is not directly interpretable (Bhati et al., 2025; Vale et al., 2022; Rudin, 2019; Huang et al., 2024; Taskin et al., 2024; Mia et al., 2023). SOM functions as an inherently interpretable white-box framework. Its prototype-based structure allows each neuron to represent a characteristic environmental state, enabling experts to trace variable relationships through visual tools such as component planes and U-matrices that reveal clusters, gradients, and transitions within the data (Wickramasinghe et al., 2021; Casalino et al., 2024; Creux et al., 2023). Such natural interpretability facilitates transparent evaluation of model behaviour and strengthens human-centred decision making in dynamic aquatic systems.

Projection-based methods such as t-SNE or UMAP can generate visually appealing two-dimensional embeddings that effectively capture local similarities among data points (Maaten and Hinton, 2008; McInnes et al., 2018). However, their stochastic behaviour, parameter sensitivity, and weaker control over topological structure reduce reproducibility and interpretability. These limitations are problematic in settings — such as environmental studies — where stable, human-readable representations are essential and where low-dimensional, topology-aware mappings (e.g., SOM-based, including hierarchical maps) are often preferable. Unlike these stochastic approaches, SOM provides a deterministic mapping that ensures reproducible visualization and stable pattern discovery, features essential for transparent environmental modelling.

Digital-transformation studies show that AI adoption in the water sector remains limited due to shortages in analytical expertise, low levels of data literacy, and lack of standardized modelling workflows (Rapp et al., 2023; Boyle et al., 2022; Grove et al., 2024). These constraints are directly influenced by the quality of decisions related to input selection, preprocessing strategies, and map configuration. The effective interpretation of component planes, U-matrices, and prototype vectors requires specialized training, and SOM outputs are not routinely integrated into operational dashboards or reporting tools. Collectively, these factors contribute to operational challenges in deploying SOM within utility settings. Despite these challenges, SOM retains several characteristics that position it as a promising candidate for applied use. Its low computational cost, transparent and interpretable structure, and deterministic behaviour make it well suited for human-in-the-loop analysis, namely compared with deep-learning models. With targeted workforce training, clearer parameterization guidelines, and integration into user-oriented interfaces, SOM could evolve from a predominantly exploratory research tool into a practical and interpretable component of routine water quality assessment workflows.

## 6. Conclusion

This systematic literature review comprehensively analysed the application of SOM for water quality assessment of lakes and reservoirs, addressing key methodological and practical dimensions. SOM has been widely employed as an exploratory tool with demonstrated strengths in clustering, visualizing, and interpreting complex, high-dimensional datasets. It has proven particularly effective for analysing a broad range of environmental parameters, especially physicochemical and nutrient indicators. The reviewed studies predominantly relied on in-situ monitoring data, supplemented with field measurements, remote-sensing products, or historical records, Datasets were characterized by various spatial and temporal sampling strategies, where more frequent sampling and denser spatial coverage generally provided enhanced analytical accuracy and deeper insights into environmental relationships. Overall, the versatility of SOM revealed hidden ecological patterns, identified environmental trends, and enabled clearer interpretation of multidimensional water quality dynamics.

Despite these strengths, the application of SOM was associated with specific limitations. These included sensitivity to parameter selection, the influence of selected SOM grid dimensions, and a general dependence on expert interpretation for meaningful analysis. The choice of SOM libraries and computational tools also varied across studies, occasionally affecting methodological consistency and reproducibility. Future research directions should include the integration of real-time data streams from IoT technologies, the exploration of multimodal or hybrid data sources, and the evaluation of alternative SOM architectures aimed at improving adaptability, interpretability, and responsiveness in dynamic freshwater environments. Although interpretative variability is inherent to SOM, this feature underscores its suitability for human-centred workflows where expert judgement remains integral to the analytical process. By enabling transparent, topology-aware exploration, SOM strengthens human-in-the-loop environmental analysis. Continued refinement and wider adoption of SOM methodologies offer meaningful potential to improve sustainable water-resource management and to provide robust exploratory tools that support informed, expert-guided decision-making.

## CRediT authorship contribution statement

**Oraib Almegdadi:** Writing – review & editing, Writing – original draft, Visualization, Validation, Methodology, Formal analysis, Data curation, Conceptualization. **João Marcelino:** Writing – review & editing, Validation, Supervision, Formal analysis. **Sarah Fakhreddine:** Writing – review & editing, Validation, Supervision, Methodology, Formal analysis. **João Manso:** Writing – review & editing, Validation, Formal analysis. **Nuno C. Marques:** Writing – review & editing, Validation, Supervision, Methodology, Formal analysis, Conceptualization.

## Declaration of Generative AI and AI-assisted technologies in the writing process

ChatGPT-4o (OpenAI) was used solely to improve the readability and language of the manuscript, in full compliance with editorial guidelines. The usage was limited to language assistance under human oversight and control; all suggestions were carefully reviewed, edited, and rewritten by the authors, who assume full responsibility for the final content.

## Funding

This work was supported by UID/04516/NOVA Laboratory for Computer Science and Informatics (NOVA LINCS), with financial support from FCT.IP.

The first author further acknowledges funding by Fundação para a Ciência e a Tecnologia (Portuguese Foundation for Science and Technology) through the Carnegie Mellon Portugal Program under the fellowship reference PRT/BD/155062/2024.

## Declaration of competing interest

The authors declare the following financial interests/personal relationships which may be considered as potential competing interests: Oraib Almegdadi reports financial support from a Carnegie Mellon Portugal (CMU-Portugal) scholarship and declares no competing interests. The remaining authors declare no known competing financial interests or personal relationships that could have appeared to influence the work reported in this paper.

## Data availability

The supplementary material and metadata extracted from the reviewed studies are openly available at https://github.com/oraibalmegdadi/Systematic-Review-SOM-in-Water-Quality-Assessment.

O. Almegdadi et al.                                                                                         Ecological Informatics 93 (2026) 103542